\documentclass[10pt,twocolumn,letterpaper]{article}

\usepackage[pagenumbers]{cvpr} 

\usepackage{graphicx}
\usepackage{amsmath}
\usepackage{amssymb}
\usepackage{booktabs}

\usepackage{multirow}
\usepackage[utf8]{inputenc} 
\usepackage[T1]{fontenc}    
\usepackage{url}            
\usepackage{amsfonts}       
\usepackage{nicefrac}       
\usepackage{microtype}      

\usepackage{algorithm}
\usepackage{algpseudocodex}

\usepackage{comment}

\graphicspath{{./figures/}}

\usepackage[pagebackref,breaklinks,colorlinks]{hyperref}

\usepackage[capitalize]{cleveref}
\crefname{section}{Sec.}{Secs.}
\Crefname{section}{Section}{Sections}
\Crefname{table}{Table}{Tables}
\crefname{table}{Tab.}{Tabs.}

\def\cvprPaperID{10} 
\def\confName{CVPR Workshop}
\def\confYear{2023}

\begin{document}

\title{A Robust Likelihood Model for Novelty Detection}

\author{Ranya Almohsen\thanks{Denotes equal contribution.} \quad
Shivang Patel$^*$ \quad
Donald A. Adjeroh \quad
Gianfranco Doretto\\
West Virginia University\\
Morgantown, WV 26506\\
{\tt\small \{ralmohse, sap00008, daadjeroh, gidoretto\}@mix.wvu.edu}
}
\maketitle

\begin{abstract}
  Current approaches to novelty or anomaly detection are based on deep neural networks. Despite their effectiveness, neural networks are also vulnerable to imperceptible deformations of the input data. This is a serious issue in critical applications, or when data alterations are generated by an adversarial attack. While this is a known problem that has been studied in recent years for the case of supervised learning, the case of novelty detection has received very limited attention. Indeed, in this latter setting the learning is typically unsupervised because outlier data is not available during training, and new approaches for this case need to be investigated. We propose a new prior that aims at learning a robust likelihood for the novelty test, as a defense against attacks. We also integrate the same prior with a state-of-the-art novelty detection approach. Because of the geometric properties of that approach, the resulting robust training is computationally very efficient. An initial evaluation of the method indicates that it is effective at improving performance with respect to the standard models in the absence and presence of attacks.
\end{abstract}



\section{Introduction} 
\label{sec:intro} 

Recognizing inliers or outliers with respect to a probability distribution is a task known as \emph{novelty} or anomaly \emph{detection}~\cite{ruff2021unifying}. It is a fundamental problem in many applications, and when computer vision supports agents exploring the world ``in-the-wild'' it is expected that sensed data might not belong to the distribution with respect to which models were previously trained. Such data has to be detected as being \emph{out of distribution} and subsequent appropriate action should be taken for its processing (e.g., open-set recognition~\cite{Scheirer_2013_TPAMI}). This novelty detection task is inherently challenging because out of distribution data is normally not available, or even dangerous to obtain in certain applications; thus, using fully unsupervised approaches is often mandatory.

Among the most succesful methods for novelty detection there are those based on deep neural networks~\cite{ruff2018deep, pidhorskyi2018generative,Perera_2019_CVPR}. Although very effective, neural networks are vulnerable to even small perturbations of the input~\cite{szegedy2014intriguing}. This means that inlier or outlier samples could be easily misclassified, despite a seemingly unnoticeable change, and it might even be possible that such changes could be operated by an adversarial entity, effectively instantiating an adversarial attack. For the case of supervised learning the problem has received considerable attention, and several approaches have been designed to deploy different types of defenses, starting from one of the most popular and effective, which is adversarial training\cite{goodfellow2015explaining}. Surprisingly, very little attention has been devoted to the development of defenses for the unsupervised case of novelty detection.

In this work, we propose a new learning prior as a defense mechanism against input data distortions or attacks that might affect a novelty detection test. We specifically focus on the case where the test statistic is the likelihood of the input data sample, which is also a very general and principled approach to novelty detection. We take a robust optimization approach, where instead of optimizing a robust supervised loss~\cite{madry2017towards}, we aim at learning a robust likelihood statistic. We then integrate this principle with a recent probabilistic novelty detection approach~\cite{almohsenKAD22cvprw}. We show that because of the geometric properties of that method, the implementation of the proposed defense prior is particularly advantageous, since the synthetic generation of outliers and inliers turns out to be very efficient. We present an initial study of the proposed approach, and show that the performance metrics improve when the robust model is compared against the starting model under standard benchmarks, as well as when the benchmarks are under attacks.


\section{Related Work}
\label{sec:relatedwork}

Novelty and anomaly detection have been studied in many domains. The central idea is to learn a model of normality from \emph{in distribution} data in an unsupervised manner such that during training no prior knowledge is available about abnormal, \emph{out of distribution} samples. Under this formulation \emph{novelty} and \emph{anomaly} detection are used interchangeably~\cite{ruff2021unifying, salehi2021unified}. Here we review several traditional and deep learning based approaches.

\textbf{Traditional Approaches.} Most traditional novelty detection approaches are based on either density estimation~\cite{eskin2000anomaly,basharat2008learning,kim2012robust} or reconstruction~\cite{abdi2010principal}. One-Class Support Vector Machines (OCSVM)~\cite{scholkopf1999support} and its extension Support Vector Data Description (SVDD)~\cite{tax2004support} are unsupervised methods, where the former learns a boundary around samples of a normal class, and the latter uses an hypersphere containing all normal samples with a minimum radius. The performance of these approaches is reported to degrade on complex high dimensional datasets~\cite{chalapathy2018anomaly}. Other unsupervised approaches include Robust Principal Component Analysis (RPCA)~\cite{candes2011robust}, and Isolation Forest (IF)~\cite{liu2008isolation}. RPCA learns a linear subspace and it identifies the anomalies in the training data, thereby removing them and retraining at each iteration.
IF tries to isolate anomalies from normal samples via successive random partitions of the feature space.
Compared to those approaches, our method learns how to compute the likelihood of data samples and does so by making the likelihood training robust.

\textbf{Deep Learning Approaches.}  Traditional approaches have been extended with deep learning. One-class Neural Network (OC-NN)~\cite{chalapathy2018anomaly} is the first approach that integrates the OC-SVM loss in the network training. Deep SVDD~\cite{ruff2018deep} instead works by jointly training a deep neural network while optimizing a data-enclosing hypersphere in the output space. Autoencoders have been used effectively for learning representations of the normal distribution~\cite{sakurada2014anomaly,zhou2017anomaly}. They learn common features in normal data and abnormal samples cannot be reconstructed accurately because they usually contain also other features, although it has been reported that different types of out of distribution samples can sometimes be reconstructed reasonably well~\cite{tong2020fixing}. Some variants of the autoencoders proposed for anomaly detection include: denoising autoencoders~\cite{vincent2008extracting}, sparse autoencoders~\cite{makhzani2013k}, variational autoencoders (VAEs)~\cite{an2015variational}, and deep convolutional autoencoders (DCAEs)~\cite{masci2011stacked, makhzani2015winner}. We also use an autoencoder, but the likelihood model that we build on, does not rely only on reconstruction error, and therefore it is less affected by the potential issues related to the reconstruction of outliers.

Other approaches combine autoencoders with Generative Adversarial Networks GANs~\cite{goodfellow2020generative}. AnoGAN~\cite{schlegl2017unsupervised} trains a GAN to generate samples according to the normal training data. At inference time given a test sample AnoGAN finds the latent representation that best reconstructs the sample. The anomaly score is based on the reconstruction error. AnoGAN is effective but not computationally efficient. Efficient GAN Based Anomaly Detection (EGBAD) addresses the performance issues of AnoGAN by adopting a Bidirectional GAN architecture~\cite{donahue2016adversarial}. In~\cite{abati2019latent} they proposed to model a latent distribution obtained from a deep autoencoder using an auto-regressive network. \cite{tuluptceva2020perceptual} leverages GANs to learn the latent distribution of normal data and uses a perceptual loss for the detection of image abnormality. Our approach also builds on an architecture that combines autoencoders with GANs under the form of adversarial autoencoders as in Generative Probabilistic Novelty Detection (GPND)~\cite{pidhorskyi2018generative}, but we build on the additional geometric properties of that architecture that were introduced in~\cite{almohsenKAD22cvprw}, and design an efficient procedure to make the likelihood training robust.

Despite their success, GAN-based approaches for anomaly detection suffer from several training issues such as mode collapse~\cite{thanh2020catastrophic}, non-convergence and instability that leads to oscillations during training, instead of a fixed-point convergence~\cite{saxena2021generative}. On the other hand, autoencoders based architectures are more stable and more convenient to train, but can overfit to a pass-through identity (null) function, and potentially reconstruct outliers when they share common features with the normal class~\cite{gong2019memorizing}. To prevent this, regularization in the form of adding deliberate perturbation to the input data often takes place. \cite{salehi2021arae} proposed the Adversarially Robust Autoencoder (ARAE), which works by forcing perceptually similar samples to be mapped closer in their latent representations. This is achieved by crafting adversarial examples that are perceptually similar to the input, but also have distant latent encoding from it. 
\cite{adey2021autoencoders} trains the autoencoder to directly output the desired per-pixel measure of abnormality without first having to perform reconstruction. This is achieved by corrupting training samples with noise and then predicting how pixels need to be shifted to remove the noise. \cite{jewell2022one} proposed the One-Class Learned Encoder-Decoder (OLED) an adversarial framework for novelty detection in both images and videos. Rather than noise perturbations a Mask Module based on a convolutional autoencoder learns to cover the most important parts of images, and the a Reconstructor is another encoder-decoder that reconstructs the masked images. \cite{barker2023robust} introduced Adversarially Learned Continuous Noise (ALCN), which is an approach to maximally globally corrupt the input prior to denoising and verified its benefits for novelty detection.

The use of perturbed data has been studied in the area of adversarial attacks. \cite{madry2017towards} provides evidence that deep neural networks for supervised learning can be made resistant to adversarial attacks. They study the adversarial robustness of neural networks in terms of robust optimization. Other methods instead are based on adding priors for regularizing the supervised loss~\cite{kurakin2017adversarial,Zhang2019-iu}. Related to novelty detection, instead, \cite{goodge2021robustness} focusses on examining the adversarial impact to deep autoencoders, and introduces a defense strategy. Similarly, \cite{Lo2023-hp} introduces Principal Latent Space, a defense strategy that is applicable to autoencoders based novelty detection approaches, and that resembles PCA-based denoising done in the latent space.

Our approach also leverages perturbed data, but with the goal of learning for the first time a likelihood function that is robust, as opposed to ``robustifying'' a supervised loss function, or proposing defense techniques that are transferrable. This should make the novelty test statistic less prone to errors in presence of perturbations within a certain set, but the approach would also help address the overfitting problem of the underlying autoencoder architecture.


\section{A Prior for Robust Novelty Detection}
\label{sec-robust}

We assume that $x$ represents a quantity of interest, e.g., an image, which can be seen as a realization of a random variable $X$, distributed according to $p_X(x)$. Due to external factors, we assume we have access only to a modified version $\bar x = x + \delta$. Here $\delta$ could model noise, or an alteration due to an adversarial attack, or the sensing of unexpected or unknown data, as it normally happens in settings in-the-wild, when data is identified as not being \emph{in distribution} (i.e., the distribution used for training the system), but in fact, it is \emph{out of distribution}, and a subsequent appropriate action needs to be taken. Therefore, a central question to answer is whether or not the sample $\bar x$ was drawn from $p_X$. In general terms, the problem can be approached by performing the test
\begin{equation}
  p_X(\bar x) = \left\{ \begin{array}{ll}
                          \ge \gamma  & \Longrightarrow \quad \mathrm{Inlier} \\
                          < \gamma  & \Longrightarrow \quad \mathrm{Outlier} \\
                        \end{array} \right.
\label{eq-ood}
\end{equation}
where $\gamma$ is a suitable application dependent threshold.

Methods that perform test~\eqref{eq-ood} almost interchangeably use names like \emph{novelty}~\cite{Perera_2019_CVPR,pidhorskyi2018generative}, \emph{anomaly}~\cite{eskin2000anomaly,Nachman2020-xv}, \emph{outlier}, or \emph{out of distribution}~\cite{Kirichenko2020-im,Serra2022-qs} detection, although subtle differences are often drawn~\cite{ruff2021unifying}. Also, in this context, $p_X$ does not really have the meaning of probability, but rather of \emph{likelihood}, which means that it depends on a particular model that was learned from training data. Out of all the possible models, we propose to seek for one that exhibits \emph{robustness} against a set $\mathcal{S}$ of predefined perturbations $\delta \in \mathcal{S}$. This means that if $x$ belongs to the set of inliers $\mathcal{X}$, then it should be that $p_X(x+\delta|x\in \mathcal{X}) \ge \gamma$, and if $x$ is an outlier, then it should be that $p_X(x+\delta|x\in \mathcal{X}^{\complement}) < \gamma$. Seeking for a robust model would allow the novelty detector to offer improved guarantees that it will be less affected by the attacks defined by the set of perturbations, which could be either intentional, or simply due to natural environmental causes.

From this discussion, we suggest that models that aim at performing test~\eqref{eq-ood} could be made \emph{robust} by including in their training a mechanism for \textsl{maximizing} the quantity
\begin{equation}
  \begin{split}
  E [ \min_{\delta \in \mathcal{S}} \; & p_X( x+\delta | x \in \mathcal{X} ) - \gamma ] + \\
  & E [ \gamma - \max_{\delta \in \mathcal{S}} p_X(x+\delta | x \in \mathcal{X}^{\complement}) ]  \; , 
\end{split}\label{eq-robust-prior}
\end{equation}
where $E[\cdot]$ denotes expectation. The approach is based on robust optimization, which has also inspired the recent adversarial training methods for supervised learning~\cite{madry2017towards}. However, rather than ``robustifying'' a classification loss function, \eqref{eq-robust-prior} tries to make the test statistic $p_X$ robust. Maximizing the first term of the \emph{robust prior}~\eqref{eq-robust-prior} aims at ensuring that the worst attacks do not turn inliers into outliers, while the second term aims at ensuring that the worst attacks do not turn outliers into inliers.

Note that in~\eqref{eq-robust-prior} the threshold $\gamma$ cancels out and it reduces to $E [ \min_{\delta \in \mathcal{S}} p_X( x+\delta | x \in \mathcal{X} ) ] - E [ \max_{\delta \in \mathcal{S}} p_X(x+\delta | x \in \mathcal{X}^{\complement}) ]$. This suggests that maximizing~\eqref{eq-robust-prior} could be achieved by turning the two terms into a single fractional term like
\begin{equation}
  \frac{E [ \max_{\delta \in \mathcal{S}} p_X(x+\delta | x \in \mathcal{X}^{\complement}) ]}{E [ \min_{\delta \in \mathcal{S}} p_X( x+\delta | x \in \mathcal{X} ) ]}
\; .
\label{eq-robust-prior-fraction}
\end{equation}
This new \emph{fractional robust prior}~\eqref{eq-robust-prior-fraction} aims for the same goals as~\eqref{eq-robust-prior} when it is \textsl{minimized}, and it could be added to training losses as a regularizer. We note that for supervised learning, adversarial training by regularization is not new~\cite{kurakin2017adversarial,Zhang2019-iu}, but to the best of our knowledge, it is new for unsupervised novelty or anomaly detection.

To solve the optimizations in the argument of the expectations in~\eqref{eq-robust-prior} and~\eqref{eq-robust-prior-fraction}, one can take a projected gradient descent (PGD) approach by implementing the iteration
\begin{equation}
  x^{t+1} = \Pi_{x+\mathcal{S}} (x^t + \alpha \nabla_x p_X(x)) \; ,
  \label{eq-pgd}
\end{equation}
where $\Pi_{x+\mathcal{S}}$ is a projection operator. On the other hand, in \S\ref{sec-robust-implementation} we show that by leveraging the properties of our network architecture it turns out that such optimizations can be solved in closed form. To keep the paper self-contained and to introduce notation, in \S\ref{sec:approach} we summarize the novelty detection approach that we build on top of, \cite{almohsenKAD22cvprw}, while in \S\ref{sec-robust-implementation} we describe how we make it robust by defining the set of perturbations $\mathcal{S}$ and enabling the training based on the prior~\eqref{eq-robust-prior-fraction}.


\section{Generative Probabilistic Novelty Detection}
\label{sec:approach}

\begin{figure*}[th!]
  \centering
  \includegraphics[width=\linewidth]{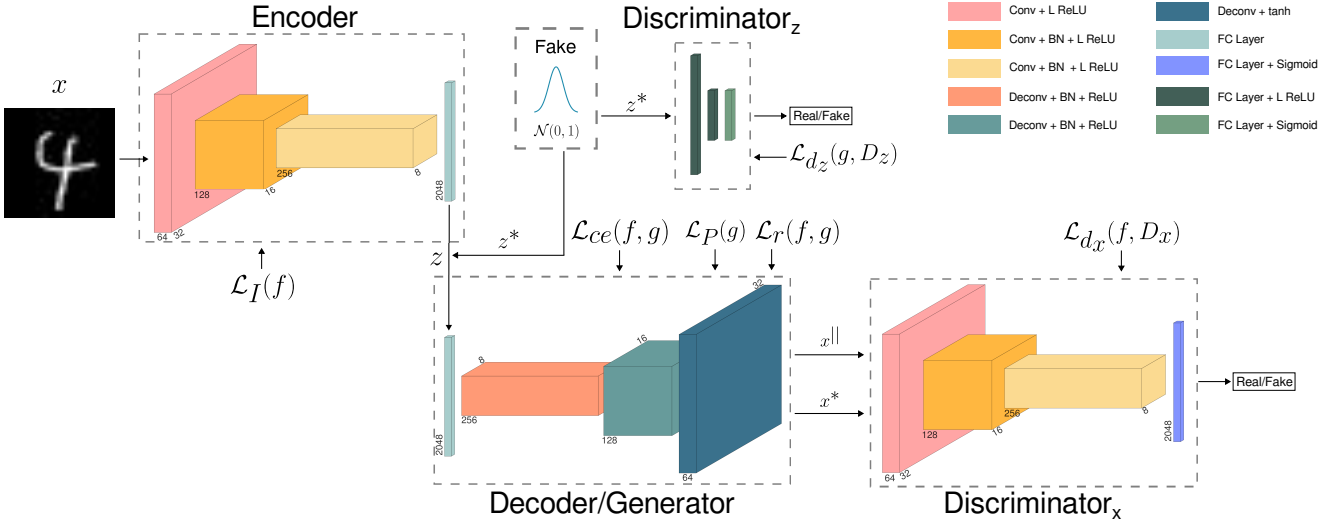}
  \caption{ \textbf{Autoencoder architecture.} Overview of the architecture and losses of the Adversarial Autoenconder (AAE)~\cite{makhzani2015adversarial} used for learning the maps $f$ and $g$. While the backbone architecture is the same as in~\cite{almohsenKAD22cvprw}, here the major addition is the use of the robust prior~\eqref{eq-robust-prior-final}. Some details of the architecture layers of the discriminators $D_x$ and $D_z$ are specified on the right. During training, the fake samples are generated from an $n$-dimensional normal distribution $\mathcal{N}(0,1)$. $x^*$ represents a mapping of $z^*$ onto the learned manifold $\mathcal{M}$, and $x^{\parallel}$ is $f(z)$.
}
  \label{fig-architecture}
\end{figure*}

We summarize the formulation, properties, and training objective function of the novelty/anomaly test initially introduced in~\cite{pidhorskyi2018generative,almohsenKAD22cvprw}. Specifically, we assume that training data points $\mathcal{D} = \{x_1, \dots, x_N \}$, where $x_i\in \mathbb{R}^m$, are sampled, possibly with noise $\xi_i$, from the model
\begin{equation}
x_i = f(z_i) + \xi_i \qquad i = 1, \cdots, N \; ,
\label{eq-model}
\end{equation}
where $z_i$ is defined in a \emph{latent} space $\Omega \subset \mathbb{R}^n$. The mapping $f:\Omega \rightarrow \mathbb{R}^m$ defines $\mathcal{M} \equiv f(\Omega)$, which is a parameterized manifold of dimension $n$, with $n<m$. It is also assumed that the Jacobi matrix of $f$ is full rank at every point of the manifold.

Given a new data point $\bar x \in \mathbb{R}^m$, the novelty test to assert whether $\bar x$ was sampled from model~(\ref{eq-model}), is derived under a number of assumptions. Specifically, $f$ is imposed to be an isometry, and in order to compute the test it is also necessary to estimate the latent representation $\bar z$ of  $\bar x$. This is done by first applying to $\bar x$ an orthogonal projection $P_{\mathcal{M}}$ from the ambient space onto $\mathcal{M}$, and then map the projection to the representation space via $f^{-1}$. This means that besides the manifold representation $f$, it is also necessary to learn a function $g$, defined as $g(x) \doteq f^{-1} \circ P_{\mathcal{M}} (x)$.

Mainly with the assumptions described above, given $\bar x$, in~\cite{pidhorskyi2018generative,almohsenKAD22cvprw} they compute the test statistic $p_X$ in~\eqref{eq-ood} as 
\begin{equation}
p_X(\bar x) = p_{Z}(\bar z) p_{X^{\perp}} (\bar x^{\perp}) \; ,
\label{eq-detection}
\end{equation}  
where $p_Z(z)$ is the probability distribution of the random variable $Z$, representing the latent space, and $\bar x^{\perp} = \bar x - f(\bar z)$, represents the component of $\bar x$ that is orthogonal to the tangent space $\mathcal{T}$ of the manifold $\mathcal{M}$. Such space is defined as $\mathcal{T} = \mathrm{span}(J_f(\bar z))$, with $J_f(\bar z)$ being the Jacobi matrix computed at $\bar z$.

The distribution $p_Z(z)$ is learned from training data by fitting a parametric generalized Gaussian distribution. Instead, the distribution $p_{X^{\perp}} (x^{\perp})$, is given by
\begin{equation}
p_{X^{\perp}} (\bar x^{\perp}) = \frac{\Gamma\left( \frac{m-n}{2}\right)}{2 \pi^{\frac{m-n}{2}} {\| \bar x^{\perp} \|}^{m-n-1}} p_{\|X^{\perp}\|} (\| \bar x^{\perp}\|) \; ,
\label{eq-p-wper}
\end{equation}
where $\Gamma(\cdot)$ is the gamma function, and $\| \cdot \|$ denotes $\ell_2$-norm. The distribution $p_{\| X^{\perp} \|} (\| x^{\perp} \|)$ is learned by computing the orthogonal projections of the training data, and histogramming the norms between data and projectons.


\subsection{Manifold Learning}

A major training task is the learning of the maps $f$, and $g$. They are modeled by an adversarial autoencoder. To satisfy the requirements, the Jacobian $J_f (z)$ will need to have orthonormal columns. This means that
\begin{equation}
  J_f(z)^{\top} J_f(z) = I \; ,
  \label{eq-f}
\end{equation} 
where $I$ is the identity matrix. Moreover, if $f$ is an isometry, then $g$ should be such that \begin{eqnarray}
\label{eq-g-1}  J_g (f(z)) J_g (f(z))^{\top} = I \; ,\\
\label{eq-g-2}  J_g (f(z)) =  J_f (z)^{\top} \; .
\end{eqnarray}
where $J_g(x)$ denotes the Jacobi matrix of $g$.

To satisfy~\eqref{eq-f}, \eqref{eq-g-1}, and \eqref{eq-g-2}, two priors are introduced.
The first is the isometry loss $\mathcal{L}_{I}(f)$, which encourages~\eqref{eq-f}, and is defined as
\begin{equation}
\mathcal{L}_{I}(f) = E \left[({\parallel} J_f(z)u{\parallel} - 1)^2 \right] \; ,
\end{equation}
where $u$ is uniformly sampled from the unit-sphere of dimension $n-1$. The second prior is the pseudo-inverse loss $\mathcal{L}_{P}(g)$, which encourages~\eqref{eq-g-1}, and is defined as
\begin{equation}
\mathcal{L}_{P}(g) = E \left[ ({\parallel}u^{\top} J_g(x) {\parallel} - 1 )^2 \right] \; ,
\end{equation}
where, again, $u$ is sampled from the same unit sphere. These priors are combined as
$\mathcal{L}_{IAE}(f,g)= \mathcal{L}_{I}(f) +\mathcal{L}_{P}(g)$.

The adversarial autoencoder architecture is shown in Figure~\ref{fig-architecture}, which follows the design in~\cite{pidhorskyi2018generative}. One adversarial component encourages the  distribution on the latent space, to be a normal distribution $\mathcal{N}(0,1)$. Another adversarial component encourages the distribution of the output of the decoder to match the distribution of real data, i.e., the manifold $\mathcal{M}$. The adversarial losses are as follows
\begin{equation}
  \small
\mathcal{L}_{d_z} (g, D_z) = E[ \log (D_z({\mathcal {N}}(0,1)))] + E[ \log (1-D_z(g(x))) ]   \; ,
\label{eq-L-dz}
\end{equation}
\begin{equation}
  \small
\mathcal{L}_{d_x} (f, D_x) = E[ \log (D_x(x))] + E[ \log (1-D_x(f(\mathcal {N}(0,1)))) ] \; ,
\label{eq-L-dx}
\end{equation}
To minimize the reconstruction error for an inlier input $x$ it is used the cross-entropy loss $\mathcal{L}_{ce} (f, g) = -E_z[ \log (p(f(g(x))|x))]$, where $\mathcal{L}_{ce}$ also encourages~\eqref{eq-g-2}. See~\cite{gropp2020isometric} for details, also on the implementation of the isometric priors above. We combine the losses that do not involve discriminators in $\mathcal{L}_{a}$
\begin{equation}
\begin{split}
  \mathcal{L}_{a}(f, g)= {\lambda}_{I} \mathcal{L}_{IAE}(f,g) + \mathcal{L}_{ce}
\end{split}
\label{eq-L-a}
\end{equation}
Where ${\lambda}_{I}$ is a balancing hyperparameter. The final objective function becomes
\begin{equation}
  \begin{split}
  \mathcal{L}(f, g, D_x, D_z) =  \mathcal{L}_{d_x} (f, D_x) + 
                                    \mathcal{L}_{d_z} (g, D_z) + \\
                                    {\lambda_a } \mathcal{L}_{a}(f, g) \; ,
                                    \end{split}
\label{eq-full}
\end{equation}
where ${\lambda_a}$ sets the trade off between the losses with and without discriminators, and $f$ and $g$ are estimated as
\begin{equation}
\hat f, \hat g = \arg \min\limits_{f, g}\,\max\limits_{D_x,D_z} \mathcal{L}(f, g, D_x, D_z) \; .
\label{eq-final}
\end{equation}


\section{Robust Likelihood Model}
\label{sec-robust-implementation}

\begin{figure*}[t!]
  \centering
  \includegraphics[width=0.8\linewidth]{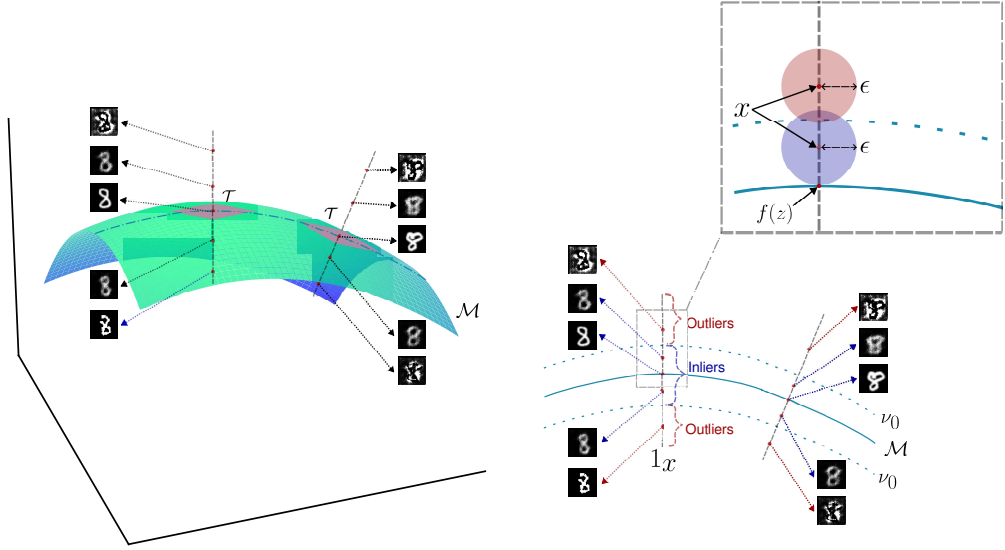}
  \vspace{-3mm}
  \caption{\textbf{Generation of inliers and outliers.} A point $x$ is an \emph{inlier} or an \emph{outlier} depending on whether its distance from the manifold $\mathcal{M}$ is below or above a threshold $\nu_0$. The orthogonal projection of $x$ onto $\mathcal{M}$ is $f(z)$. $x$ can be perturbed within a radius $\epsilon$. The strongest perturbation for an inlier/outlier occurs in the outward/inward direction orthogonal to the tangent plane $\mathcal{T}$.}
  \label{fig-sample-generation}
\end{figure*}
We now describe how we make the novelty detection method in \S\ref{sec:approach} robust, based on the ideas described in \S\ref{sec-robust}. First, we assume that the set of admissible perturbations is an $\epsilon$-ball, which means that $\mathcal{S} = \{ \delta : \| \delta \| \le \epsilon \}$. Next, we make the following simplifying assumptions
\begin{equation}
  \min_{\| \delta \| \le \epsilon} p_X(x + \delta) \approx p_X (x + \epsilon 1_x) \; ,
  \label{eq-min}
\end{equation}
\begin{equation}
  \max_{\| \delta \| \le \epsilon} p_X(x + \delta) \approx p_X (x - \epsilon 1_x) \; ,
  \label{eq-max}
\end{equation}
where $1_x = (x - f(z))/ \| x - f(z) \|$ is a unit norm vector. Given the properties of the autoencoder defined by $g$ and $f$, $f(z)$ is the orthogonal projection of $x$ onto the manifold $\mathcal{M}$, and $1_x$ is perpendicular to the tangent plane $\mathcal{T}$. Therefore, \eqref{eq-min} stems from the fact that it is reasonable to expect that the largest drop of the likelihood $p_X$ will be due to a perturbation that moves $x$ away from $\mathcal{M}$ the furthest possible, and this should happen along the direction orthogonal to $\mathcal{M}$. Similarly, \eqref{eq-max} stems from the fact that we are expecting to observe the highest increase of $p_X$ when $x$ moves the closest towards $\mathcal{M}$. See Figure~\ref{fig-sample-generation}.

We stress that \eqref{eq-min} and \eqref{eq-max} are possible thanks to the properties of the autoencoder $g\circ f$. They provide a remarkable computational saving in that the two optimizations are solved in closed-form without requiring the use of PGD~\eqref{eq-pgd}. Moreover, \eqref{eq-min} and \eqref{eq-max} also suggest a very efficient strategy for generating inliers and outliers data for training purposes, as explained below, which again does not require PGD.

Since data points are modeled according to~\eqref{eq-model}, then we have that $x = f(z) + \nu 1_x$, where $\nu \doteq \| \xi \|$. Therefore, given~\eqref{eq-min}, as a general recipe for \emph{generating inliers} from $x$, we use the following expression
\begin{equation}
  f(z) \pm (\nu + \epsilon) 1_x \; ,
  \label{eq-inliers}
\end{equation}
where the $\pm$ sign takes into account that inliers can be generated on both sides of the tangent plane $\mathcal{T}$. Similarly, given~\eqref{eq-max}, the general recipe for \emph{generating outliers} from $x$ becomes
\begin{equation}
  f(z) \pm (\nu - \epsilon) 1_x \; ,
  \label{eq-outliers}
\end{equation}
where the $\pm$ sign is introduced for the same reason as in~\eqref{eq-inliers}. Figure~\ref{fig-sample-generation} illustrates the generation process.

Note, however, that~\eqref{eq-inliers} should be used only if $f(z) \pm \nu 1_x \in \mathcal{X}$, which means it is an inlier. Similarly, \eqref{eq-outliers} should be used only if $f(z) \pm \nu 1_x \in \mathcal{X}^{\complement}$, which means it is an outlier. Deciding whether $f(z) \pm \nu 1_x$ belongs to $\mathcal{X}$ or $\mathcal{X}^{\complement}$ is straightforward once we know the value $\nu_0$ such that $p_X(f(z) \pm \nu_0 1_x) = \gamma$. From~\eqref{eq-detection} and ~\eqref{eq-p-wper}, $\nu_0$ can be computed by solving numerically the equation
\begin{equation}
  p_Z(z) \frac{\Gamma(\frac{m-n}{2})}{2 \pi^{\frac{m-n}{2}} \nu_0^{m-n-1}} p_{\| X \|^{\perp}}( \nu_0 ) = \gamma \; .
  \label{eq-boundary}
\end{equation}
It follows that $f(z) \pm \nu 1_x \in \mathcal{X}$ if $\nu \le \nu_0$, and that $f(z) \pm \nu 1_x \in \mathcal{X}^{\complement}$ if $\nu > \nu_0$. See Figure~\ref{fig-sample-generation}.

\subsection{Robust Prior}

From the previous discussion, we note that given a training dataset composed of only inliers, we can still generate synthetic outliers according to how we choose $\nu$. In particular, we propose to randomly sample inliers by assuming that $\nu \sim \mathcal{U}([0, \nu_0))$, which means $\nu$ is uniformly distributed in the interval $[0, \nu_0)$. Therefore, inliers come from the region closer to $\mathcal{M}$. Outliers instead, are sampled by assuming that $\nu \sim \mathcal{E}(\lambda)$, which means that $\nu$ is exponentially distributed with rate parameter $\lambda$, and an offset $\nu_0$ is also added.

In essence, we propose to add to the objective function~\eqref{eq-full} the following robust prior
\begin{equation}
\mathcal{L}_{r}(f, g) = \frac{E_{x \sim \mathcal{D}}[ E_{\nu \sim \mathcal{E}} [ p_X(f(z) \pm (\nu +\nu_0 - \epsilon) 1_x ) ]]}{E_{x \sim \mathcal{D}}[ E_{\mathcal{\nu \sim U}} [ p_X( f(z) \pm (\nu + \epsilon) 1_x  ) ]]}
\; .
\label{eq-robust-prior-final}
\end{equation}
The final procedure for the training of the \emph{robust likelihood novelty detection (RLND)} model is summarized in Algorithm~\ref{alg-rnd}.

\begin{algorithm}
\caption{Robust Likelihood Novelty Detection}\label{alg-rnd}
\begin{algorithmic}
  \State \textbf{Input:} Training dataset $\mathcal{D} = \{ x_i \}_{i=1}^N$.
  \State \textbf{Parameters:} Minibatch size $M$; radius $\epsilon$; rate $\lambda$
  \State Train $f$ and $g$ and obtain initial weights from~\eqref{eq-final}
  \State Obtain $\gamma$ from the validation dataset
\Repeat
\State Sample a minibatch $\{ x_{i_j} \}_{j=1}^M$
\For{$j \in \{1, \dots, M \}$}
\State Compute $\nu_{0,i_j}$ by solving \eqref{eq-boundary}
\State $\nu \leftarrow$ draw from $\mathcal{U}([0, \nu_{0,i_j}))$
\LComment{Inlier generation; randomly pick $+$ or $-$}
\State $x_j^I \leftarrow  f(z_{i_j}) \pm (\nu + \epsilon) 1_{x_{i_j}}$
\State $\nu \leftarrow$ draw from $\mathcal{E}(\lambda)$
\LComment{Outlier generation; randomly pick $+$ or $-$}
\State $x_j^O \leftarrow  f(z_{i_j}) \pm (\nu + \nu_{0,i_j}- \epsilon) 1_{x_{i_j}}$
\EndFor
\LComment{Use the inlier batch $\{ x_j^I \}$ for the following}
\State Maximize $\mathcal{L}_{d_x}$~\eqref{eq-L-dx} by updating weights of $D_x$
\State Minimize $\mathcal{L}_{d_x}$~\eqref{eq-L-dx} by updating weights of $f$
\State Maximize $\mathcal{L}_{d_z}$~\eqref{eq-L-dz} by updating weights of $D_z$
\LComment{Use the inlier batch $\{ x_j^I \}$ and outlier batch $\{ x_j^O \}$ for the following}
\State Minimize $\mathcal{L}_{a}$~\eqref{eq-L-a}, $\mathcal{L}_{d_z}$~\eqref{eq-L-dx}, and $\mathcal{L}_{r}$~\eqref{eq-robust-prior-final}, by updating weights of $g$ and $f$.
\LComment{Note that only  $\mathcal{L}_{r}$ uses both inlier and outlier batches}
\Until{Convergence}
\end{algorithmic}
\end{algorithm}


\section{Experiments}
\label{sec:experiments}

We present the evaluation of the proposed robust likelihood novelty detection (RLND) method. We compare RLND with state-of-the-art methods by using common benchmark datasets for the unsupervised novelty detection task, and we follow the same protocol as in~\cite{almohsenKAD22cvprw} to maintain consistency across all experiments. We utilize two key metrics: the $F_1$ score and the area under the ROC (AUROC). These metrics provide a comprehensive assessment of the performance of our approach.

In each experiment, the datasets are partitioned into training, validation, and testing sets using a random split. Specifically, we allocate 60\% of the data for training, where instances from each class are randomly sampled, and 20\% for validation. The remaining 20\% are reserved for testing.

\subsection{Datasets}

We utilize three benchmark datasets commonly used for novelty and anomaly detection, namely MNIST, Fashion-MNIST, and Coil-100.

\noindent \textbf{MNIST}~\cite{lecun1998mnist} is composed of 70,000 $28 \times 28$ handwritten single digits from 0 to 9.

\noindent \textbf{Fashion-MNIST}~\cite{xiao2017fashion}, similar to MNIST, contains 70,000 $28\times 28$ grayscale images of 10 fashion product categories.

\noindent \textbf{Coil-100}~\cite{nene1996columbia} is a dataset of 7,200 color images with 100 object classes. For each of 100 objects, pictures were taken in different poses, 5 degrees apart from one another, resulting in 72 images for each object.\\

\subsection{Implementation Details}
\label{sec:param-comp}

We implemented Algorithm~\ref{alg-rnd}, where the first step trains a GPNDI model. We refer to~\cite{almohsenKAD22cvprw} for picking the parameters $\lambda_I$ and $\lambda_a$. Instead, when also $\mathcal{L}_r$ is minimized, we weight it by a hyperparameter $\lambda_r$, which is set to $0.001$. In all the experiments the latent space size $n$, is set to $16$, since it has been reported to yield the highest $F_1$ score on the validation sets~\cite{pidhorskyi2018generative,almohsenKAD22cvprw}.

The initial GPNDI model is then further trained for $30$ more epochs using an NVIDIA RTX A6000 GPU and the ADAM optimizer. For all datasets we use a batch size $M=128$. Instead, to ensure optimal convergence, the learning rates are set to $0.0002$ for both MNIST and Fashion-MNIST, while for COIL-100, the learning rate is set to $0.0003$.

In Algorithm~\ref{alg-rnd}, we set the rate parameter $\lambda$ to $5.0$ in all the experiments, while the radius $\epsilon$ varies with the dataset. Specifically, the training $\epsilon$ is set to $\epsilon = 0.5 \times \nu_0$, where $\nu_0$ here is intended as averaged over the inlier training samples. This choice ensures that a sample residing at the same distance from the manifold and the decision boundary will remain inside the inliers set $\mathcal{X}$ even after the largest admissible perturbation. For MNIST and Fashion-MNIST, we used $\epsilon$ values that were averaged over all choices of inlier manifolds, and they are $2.4$ and $3.0$, respectively. For COIL-100, $\epsilon$ was determined based on the random selection of inliers. This choice of $\epsilon$ tailors the model to the specific inlier manifold being learned.

\subsection{Results without Attacks}
\label{sec:res-no-attacks}

\newcommand{\f}[1]{\textbf{#1}}
\newcommand{\s}[1]{\underline{#1}}

\begin{table}
  \caption{$F_1$ scores on MNIST~\cite{lecun1998mnist}. Inliers are taken to be images of one category, and outliers are randomly chosen from other categories. All results are averages from a 5-fold cross validation.  }
  \label{mnist-table}
  \centering
  \resizebox{1.0\columnwidth}{!}{%
  \begin{tabular}  {lllllllllll}
    \toprule
    \% of outliers &
                     $\mathcal{D(R(}X))$~\cite{sabokrou2018adversarially}    & $\mathcal{D(}X)$~\cite{sabokrou2018adversarially}      & LOF~\cite{breunig2000lof}  & DRAE~\cite{xia2015learning} & GPND~\cite{pidhorskyi2018generative} & GPNDI~\cite{almohsenKAD22cvprw}& \textbf{RLND} (Ours) \\
    
\midrule
    10 & 0.97  & $0.93$  & $0.92$  & $0.95$ & $0.983$ & \s{0.984} & \f{0.990} \\
    20 & 0.92  & $0.90$  & $0.83$  & $0.91$ & $0.971$ & \s{0.976} & \f{0.986} \\
\midrule
    30 & 0.92  & $0.87$  & $0.72$  & $0.88$ & $0.961$ & \s{0.968} &  \f{0.980} \\
    40 & 0.91  & $0.84$  & $0.65$  & $0.82$ & $0.950$ & \s{0.960} &  \f{0.977} \\
\midrule
    50 & 0.88  & $0.82$  & $0.55$  & $0.73$ & $0.939$  & \s{0.953} & \f{0.974} \\   \bottomrule
  \end{tabular}}
\end{table}




\begin{table}
  \caption{Results on Fashion-MNIST~\cite{xiao2017fashion}. $F_1$ scores where inliers are taken to be images of one category, and outliers are randomly chosen from other categories. }
  \label{mnistf}
  \centering
  \scalebox{0.8}{%
  \begin{tabular}  {llllll}
    \toprule
    \% of outliers & $10$   & $20$     & $30$  & $40$ & $50$ \\
    
\midrule

    GPND\cite{pidhorskyi2018generative} & $0.968$  & $0.945$  & $0.917$  & $0.891$ & $0.864$ \\
\midrule
    GPNDI\cite{almohsenKAD22cvprw} & $0.974$ & $0.953$  & $0.930$  & $0.904$ & $0.873$ \\
\midrule
        \textbf{RLND} (Ours) & \f{0.986}  & \f{0.977}  & \f{0.970}  & \f{0.961} & \f{0.954} \\
   \bottomrule
  \end{tabular}
 }
\end{table}


\begin{table*}
  \caption{Results on Coil-100. Inliers are taken to be images of one, four, or seven randomly chosen categories, and outliers are randomly chosen from other categories (at most one from each category).}
  \label{coil-table}
  \centering
  \resizebox{2.0\columnwidth}{!}{%

  \begin{tabular}  {lllllllllllll}
    \toprule
    & OutRank~\cite{moonesignhe2006outlier,moonesinghe2008outrank}    & CoP~\cite{rahmani2016coherence}      & REAPER~\cite{lerman2015robust}  & OutlierPursuit~\cite{xu2010robust}  & LRR~\cite{liu2010robust}  & DPCP~\cite{tsakiris2015dual} & $\ell_1$ thresholding~\cite{soltanolkotabi2012geometric} & R-graph~\cite{you2017provable} & GPND~\cite{pidhorskyi2018generative} & GPNDI~\cite{almohsenKAD22cvprw} &  \textbf {RLND} (Ours) \\
    
    \midrule
         \multicolumn{9}{c}{Inliers: \textbf{one} category of images , Outliers: $50\%$}                   \\
    \midrule
\midrule
    AUROC & $0.836$  & $0.843$ & $0.900$  & $0.908$ & $0.847$ & $0.900$ & \s{0.991} & \f{0.997} & $0.968$ & $0.984$& $0.990$ \\
    $F_1$     & $0.862$ & $0.866$  & $0.892$ & $0.902$ & $0.872$ & $0.882$ & $0.978$ & \f{0.990} & 0.979   & $0.894$ & \s{0.989}  \\
    \midrule
    \midrule
         \multicolumn{9}{c}{Inliers: \textbf{four} category of images , Outliers: $25\%$}                   \\
\midrule
\midrule
    AUROC & $0.613$  & $0.628$ & $0.877$  & $0.837$ & $0.687$ & $0.859$ & \s{0.992} & \f{0.996} &  $0.945$ & $0.960$& $0.980$  \\
    $F_1$     & $0.491$ & $0.500$  & $0.703$  & $0.686$ & $$0.541 & $0.684$ & $0.941$ & \f{0.970} &  \s{0.960} &  $0.953$ & \f{0.970} \\
       \midrule
       \midrule
    
       \multicolumn{9}{c}{Inliers: \textbf{seven} category of images , Outliers: $15\%$}                   \\
\midrule
\midrule
    AUROC & $0.570$  & $0.580$ & $0.824$ & $0.822$ & $0.628$ & $0.804$ & \s{0.991} & \f{0.996} & $0.919$ &  $0.950$ & $0.985$   \\
    $F_1$     & $0.342$ & $0.346$  & $0.541$& $0.528$ & $0.366$ & $0.511$ & $0.897$ & $0.955$ & 0.941  &\s{0.964} & \f{0.981} \\   \bottomrule
  \end{tabular}}
\end{table*}


\begin{table}
  \caption{Precision, Recall, $F_1$ and AUROC measures for various $\epsilon$-attacks on the MNIST test set.}
  \label{attack-table}
  \centering
  \scalebox{0.62}{%
  \begin{tabular}  {llllll|lllll}
    \toprule
    & \multicolumn{5}{c}{GPNDI} & \multicolumn{5}{c}{\textbf{RLND} (Ours)}\\
    $\epsilon$ & 0.0 & 0.5   & 1.0    & 2.0  & 3.0 & 0.0 &  0.5   & 1.0 & 2.0 & 3.0\\
    \cmidrule(l{2pt}r{2pt}){2-11}
    Precision & 0.971 & 0.946 & 0.885 & 0.704 & 0.576 & 0.971 & 0.954 & 0.924 & 0.745 & 0.605 \\
    Recall    & 0.951 & 0.925 & 0.902 & 0.905 & 0.950 & 0.977 & 0.946 & 0.915 & 0.908 & 0.950 \\
    $F_1$        & 0.961 & 0.935 & 0.893 & 0.781 & 0.706 & 0.974 & 0.950 & 0.919 & 0.806 & 0.726 \\
    \cmidrule(l{2pt}r{2pt}){1-11}
    AUROC       & 0.99 & 0.979 & 0.948 & 0.781 & 0.470 & 0.993 & 0.985 & 0.966 & 0.850 & 0.581\\
   \bottomrule
  \end{tabular}}
\end{table}

\begin{table}
  \caption{Precision, Recall, $F_1$ and AUROC measures for various $\epsilon$-attacks on the Fashion-MNIST test set.}
  \label{FMNIST-attack-table} 
  \centering
  \scalebox{0.62}{%
  \begin{tabular}  {llllll|lllllll}
    \toprule
    & \multicolumn{5}{c}{GPNDI} & \multicolumn{5}{c}{\textbf{RLND} (Ours)}\\
    $\epsilon$ & 0.0 & 0.5   & 1.0    & 2.0  & 3.0 & 0.0 &  0.5   & 1.0 & 2.0 & 3.0\\
    \cmidrule(l{2pt}r{2pt}){2-11}
    Precision & 0.939 &  0.905 & 0.856 & 0.741 & 0.608 & 0.954 & 0.932 & 0.912 & 0.840 & 0.744 \\
    Recall    & 0.931 &  0.931 & 0.924 & 0.909 & 0.958 & 0.956 & 0.947 & 0.934 & 0.915 & 0.917 \\
    $F_1$        & 0.937 &  0.921 & 0.886 & 0.811 & 0.737 & 0.954 & 0.939 & 0.922 & 0.881 & 0.814 \\
    \cmidrule(l{2pt}r{2pt}){1-11}
    AUROC       & 0.979 &  0.921 & 0.935 & 0.834 & 0.647 & 0.987 & 0.980 & 0.968 & 0.923 & 0.836 \\
   \bottomrule
  \end{tabular}}
\end{table}

\begin{figure}
  \centering
  \includegraphics[width=.8\linewidth]{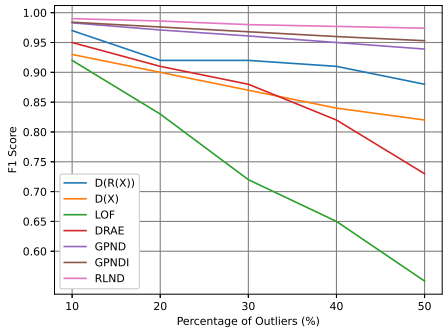}
  \vspace{-3mm}
  \caption{ Results on MNIST~\cite{lecun1998mnist} dataset.}
  \vspace{-3mm}
  \label{fig-mnist}
\end{figure}

\noindent \textbf{MNIST dataset.} For the MNIST dataset, we compose five random balaced data splits to evaluate the performance of our approach. We use three splits for training, reserving one split for validation and one for testing. The value of $\gamma$ that yields the highest $F_1$ score on the validation set is then employed during the testing phase. We designate each digit as an inlier, while the remaining digit samples are selected to generate outlier percentages ranging from 10\% to 50\%. This allows us to assess the robustness of our approach across different levels of novelty in the dataset. We present these results in Table~\ref{mnist-table} and illustrate them graphically in Figure~\ref{fig-mnist}. The comparative evaluation against GPND, GPNDI and other methods, highlights that a consistent improvement is achieved. This suggests that the additional training based on the robust prior~\eqref{eq-robust-prior-final}, which is the foundation of RLND, leads to better performance on this standard benchmark.

\begin{figure}
  \centering
  \includegraphics[width=.8\linewidth]{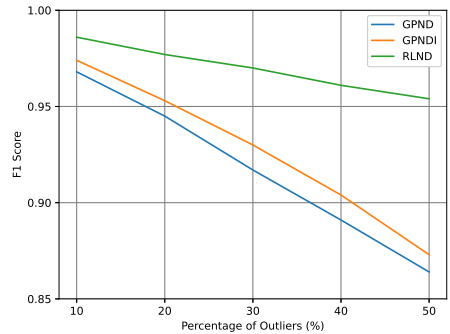}
  \vspace{-3mm}
  \caption{ Results on Fashion-MNIST dataset.}
  \vspace{-3mm}
  \label{fig-fmnist}
\end{figure}

\noindent \textbf{Fashion-MNIST dataset.} We maintain consistency with the protocol followed for the MNIST dataset when conducting experiments on Fashion-MNIST. The results of these experiments are presented in Table~\ref{mnistf}, and visually depicted in Figure~\ref{fig-fmnist}. It can be seen that the robust training of RLND, although designed for a specific class of perturbations, it can lead to a remarkable increase in performance even when data is not necessarily undergoing the same type of perturbations. This result further validates the effectiveness of RLND.

\noindent \textbf{Coil-100 dataset.} Similar to previous datasets, we adopt a 5-fold cross-validation approach for evaluating the performance on the Coil-100 dataset. However, in this case, we utilize four splits for training and reserve one split for testing. The optimal value of $\gamma$ is determined based on the training set, ensuring the best possible performance during testing. For each experiment, we randomly select 1, 4, or 7 classes as inliers, while considering the remaining classes as outliers. The outliers are included at percentages of 50\%, 25\%, and 15\%, respectively. The results are presented in Table~\ref{coil-table}. Our RLND approach consistently outperforms GPNDI in all cases, further confirming its superior performance, and the usefulness of a robust approach. Furthermore, we note that RLND is able to match or surpass the $F_1$ scores of R-graph~\cite{you2017provable}. This is significant because R-graph is based on a large pre-trained VGG network, whereas we are training from scratch a very small autoencoder architecture with a limited number of samples, which is around $70$ per class.

\subsection{Results with Attacks}

In this experiment, our primary objective is to evaluate the robustness of our model against $\epsilon$-attacks. $\epsilon$-attacks involve perturbing the input data point $x$ along the $1_x$ direction from the manifold's projection. We conduct this test on the MNIST and Fashion-MNIST datasets with 50\% of outliers. The attack on a testing inlier sample $x$ is generated by adding a perturbation $\epsilon 1_x$. The attack on a testing outlier sample $x$ is generated by subtracting a perturbation $\epsilon 1_x$.

To assess the model's performance under attack, we measure precision, recall, $F_1$ score, and AUROC with varying values of $\epsilon$. For both datasets, GPNDI and RLND were trained as described in \S\ref{sec:param-comp} and \S\ref{sec:res-no-attacks}. In particular, for a given dataset, the training $\epsilon$ value is the same for every $\epsilon$-attack. The results are reported in Table~\ref{attack-table} and Table~\ref{FMNIST-attack-table}. We note that RLND outperforms GPNDI according to all the metrics, conditions, in both datasets, and by a significant margin. This is a very encouraging result, which supports the major contribution of the proposed approach. We further note the quick and strong deterioration in performance of the baseline approach GPNDI, as $\epsilon$ grows, which is clearly due to the fact that GPNDI was not robustly trained to respond to these attacks. RLND instead, demonstrates a much smaller rate of deterioration.


\section{Conclusion}
\label{sec:conclusion}

In this work we introduced a new prior for learning a likelihood model for novelty or anomaly detection that is robust to a predefined set of perturbations. We then integrated this prior with GPNDI, an existing method for novelty detection, which is based on computing the likelihood of the input samples. The integration, referred to as Robust Likelihood Novelty Detection (RLND), is computationally efficient, and entails a training refinement of the initial model, by optimizing an updated loss with minibatches of sampled synthetically generated inliers and outliers. Our initial results reveal that integrating the robust prior leads to a clear performance improvement over the baseline method, when both are tested on the benchmark datasets MNIST, Fashion-MNIST, and COIL-100. This means that the prior is a beneficial regularizer when perturbations, or attacks are not present. Furthermore, when both the baseline model, GPNDI, and the robust model, RLND, are subject to $\epsilon$-attacks, we observed that the proposed method can cope with the attacks significantly better than the baseline. While these are very encouraging results, in future work we plan to address other areas of investigation that we currently left out, such as testing our approach against other types of adversarial attacks, like those based on PGD.


\section*{Acknowledgments}

This material is based upon work supported by the National Science Foundation under Grants No. 1920920, 2125872 and 2223793.

{\small
\bibliographystyle{ieee_fullname}
\bibliography{main-references}
}

\end{document}